\documentclass{bmvc2k}

\usepackage{graphicx}
\usepackage{tikz}
\usepackage{comment}
\usepackage{amsmath,amssymb} % define this before the line numbering.
\usepackage{color}
\usepackage{wrapfig}

\usepackage[accsupp]{axessibility}  % Improves PDF readability for those with disabilities.
\usepackage{xcolor}
\usepackage{epsfig}
\usepackage{bm}
\usepackage{gensymb}
\usepackage{eucal}
\usepackage{mathrsfs}
\usepackage{makecell}
\usepackage{multirow, booktabs}
\usepackage{enumitem,kantlipsum}
\usepackage{ulem}

\def\bs{\bm}

\title{APSNet: Attention Based Point Cloud Sampling}

\addauthor{Yang Ye}{yye10@student.gsu.edu}{1}
\addauthor{Xiulong Yang}{xyang22@student.gsu.edu}{1}
\addauthor{Shihao Ji}{sji@gsu.edu}{1}

\addinstitution{
 Department of Computer Science\\
 Georgia State University\\
 Atlanta, US
}

\runninghead{Yang Ye, Xiulong Yang, and Shihao Ji}{APSNET}

\begin{document}

\maketitle

\begin{abstract}
Processing large point clouds is a challenging task. Therefore, the data is often downsampled to a smaller size such that it can be stored, transmitted and processed more efficiently without incurring significant performance degradation. Traditional task-agnostic sampling methods, such as farthest point sampling (FPS), do not consider downstream tasks when sampling point clouds, and thus non-informative points to the tasks are often sampled. This paper explores a task-oriented sampling for 3D point clouds, and aims to sample a subset of points that are tailored specifically to a downstream task of interest. Similar to FPS, we assume that point to be sampled next should depend heavily on the points that have already been sampled. We thus formulate point cloud sampling as a sequential generation process, and develop an attention-based point cloud sampling network (APSNet) to tackle this problem. At each time step, APSNet attends to all the points in a cloud by utilizing the history of previously sampled points, and samples the most informative one. Both supervised learning and knowledge distillation-based self-supervised learning of APSNet are proposed. Moreover, joint training of APSNet over multiple sample sizes is investigated, leading to a single APSNet that can generate arbitrary length of samples with prominent performances. Extensive experiments demonstrate the superior performance of APSNet against state-of-the-arts in various downstream tasks, including 3D point cloud classification, reconstruction, and registration. Our code is available at \url{https://github.com/Yangyeeee/APSNet}.
\end{abstract}

\vspace{-5pt}
\section{Introduction}\vspace{-5pt}
 With the rapid development of 3D sensing devices (e.g., LiDAR and RGB-D camera), huge point cloud data are generated in the areas of robotics, autonomous driving and virtual reality~\cite{Nchter2008TowardsSM,KITTI,AR}. A 3D point cloud, composed of the raw coordinates of scanned points in a 3D space, is an accurate representation of an object or shape and plays a key role in perception of the surrounding environment. Since point clouds lie in irregular space with variable densities, traditional feature extraction methods, such as convolutional neural networks (CNNs), designed for grid-structured 2D data do not perform well on 3D point clouds. Some methods attempt to first stiffly transform point clouds into grid-structured data and then take advantage of CNNs for feature extraction, such as projection-based methods~\cite{Simony_2018_ECCV_Workshops,8569311} and volumetric convolution-based methods~\cite{7989161,8205955}. Because placing a point cloud on a regular grid generates an uneven number of points in grid cells, applying the same convolution operation on such grid cells leads to information loss in crowded cells and wasting computation in empty cells. Recently, many methods of directly processing point cloud~\cite{qi2017PointNetplusplus, yu2018pu-net, li2018pointcnn, qi2019deep} have been proposed to enable efficient computation and performances in many applications, such as 3D point cloud classification~\cite{qi2017PointNetplusplus, li2018pointcnn, thomas2019kpconv, wu2019pointconv}, semantic segmentation ~\cite{li2018so-net, su2018splatnet, liu2019relation, wang2019dynamic,wang2018sgpn} and reconstruction~\cite{achlioptas2018learning, yang2018folding, han2019multi, zhao20193dpoint}, have been improved significantly. These methods take raw point clouds as input (without quantization) and aggregate local features at the last stage of the network, so the accurate data locations are kept intact but the computation cost grows linearly with the number of points. However, processing large-scale dense 3D point clouds is still challenging due to the high cost of storing, transmitting and processing these data. Point cloud sampling, a task of selecting a subset of points to represent the original point clouds at a sparse scale, can reduce data redundancy and improve the efficiency of 3D data processing. So far, there are a few heuristics-based sampling methods, such as random sampling (RS), farthest point sampling (FPS)~\cite{eldar1997FPS, moenning2003fast}, and grid (voxel) sampling~\cite{wu2015modelnet, qi2016volumetric}. However, all of these methods are task-agnostic as they do not take into account the subsequent processing of the sampled points and may select non-informative points to the downstream tasks, leading to suboptimal performance. Recently, S-NET~\cite{dovrat2019learning_to_sample} and SampleNet~\cite{lang2020SampleNet} are proposed, which demonstrate that better sampling strategies can be learnt via a task-oriented sampling network. These sampling networks can generate a small number of samples that optimize the performance of a downstream task, and outperform traditional task-agnostic samplers significantly in various applications~\cite{dovrat2019learning_to_sample, lang2020SampleNet}.

%The deep neural network will be optimized with the task loss from the backbone network and Chamfer Distance loss to regulate the distribution of sampled points.

%In contrast to the existing methods, 

%landrieu2018large

\begin{figure*}[t]%\vspace{10pt}
\centerline{
\includegraphics[width = 12cm]{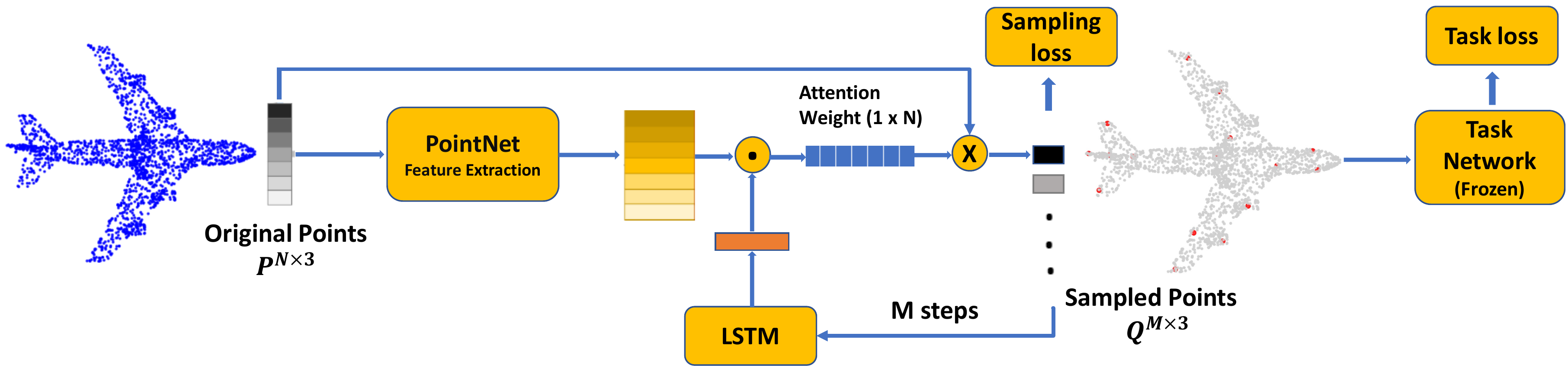}}\vspace{-5pt}
\caption{Overview of APSNet. APSNet first extracts features with a simplified PointNet that preserves the geometric information of a point cloud. Then, an LSTM with attention mechanism is used to capture the relationship among points and select the most informative point sequentially. Finally, the sampled point cloud is fed to a task network for prediction. The whole pipeline is optimized by minimizing a task loss and a sampling loss jointly.}
\label{fig:overview}
\vspace{-15px}
\end{figure*}

We argue that point cloud sampling can be considered as a sequential generation process, in which points to be sampled next should depend on the points that have already been sampled. However, existing task-oriented sampling methods, such as S-NET~\cite{dovrat2019learning_to_sample} and SampleNet~\cite{lang2020SampleNet}, do not pay enough attention to the sample dependency and generate all samples in one shot (without parameter reusing or sharing when generating samples of different sizes). In this paper, we propose an attention-based point cloud sampling network (APSNet) for task-oriented sampling, which enables a FPS-like sequential sampling but with a task-oriented objective. Specifically, APSNet employs a novel LSTM-based sequential model to capture the correlation of points with a global attention. The feature of each point is extracted by a simplified PointNet architecture, followed by an LSTM~\cite{lstm} with attention mechanism to capture the relationship of points and select the most informative ones for sampling. Finally, the sampled point cloud is fed to a (frozen) task network for prediction. The whole pipeline is fully differentiable and the parameters of APSNet can be trained by optimizing a task loss and a sampling loss jointly (See Fig.~\ref{fig:overview}). 

Depending on the availability of labeled training data, APSNet can be trained in supervised learning or self-supervised learning via knowledge-distillation~\cite{kd14}. In the latter case, no ground truth label is needed for the training of APSNet. Instead, the soft predictions of task network are leveraged to train APSNet. Interestingly, the self-supervised training of APSNet achieves impressive results that are close to the performance of supervised training. This makes APSNet widely applicable in situations where only a deployed task net is available but the original labeled training dataset of the task net is no longer accessible. 

In addition, given the autoregressive model of APSNet, our method can generate arbitrary length of samples from a single model. This entails an effective joint training of APSNet with multiple sample sizes, resulting in a single compact model for point cloud sampling, while S-NET and SampleNet require a growing model size to generate larger sized point samples and the parameter reusing or sharing is not as effective as APSNet~\cite{dovrat2019learning_to_sample,lang2020SampleNet}. Our main contributions are summarized as follows:
\begin{enumerate}[leftmargin=*, topsep=1pt, partopsep=1pt, itemsep=1pt,parsep=1pt]
\item  We propose APSNet, a novel attention-based point cloud sampling network, which enables a FPS-like sequential sampling with a task-oriented objective.
\item  APSNet can be trained in supervised learning or self-supervised learning via knowledge-distillation, while the latter requires no ground truth labels for training and is thus widely applicable in situations where only a deployed task network is available.
\item APSNet can be jointly trained with multiple sample sizes, yielding a single compact model that can generate arbitrary length of samples with prominent performance.
\item Compared with state-of-the-art sampling methods, APSNet demonstrate superior performance on various 3D point cloud applications.
\end{enumerate}

\vspace{-5pt}
\section{Related Work}\vspace{-5pt}

\paragraph{Deep Learning on Point Clouds}
%These representations enable the natural extension of successful neural processing paradigms from 2D images to 3D data. 
% transfer point cloud to occupancy grid and
Following the breakthrough results of CNNs in 2D image processing tasks~\cite{AlexNet,he2016deep}, there has been a strong interest of adapting such methods to 3D geometric data. Compared to 2D images, point clouds are sparse, unordered and locality-sensitive, making it non-trivial to adapt CNNs to point cloud processing. Early attempts focus on regular representations of the data in the form of 3D voxels~\cite{wu2015modelnet, qi2016volumetric}. These methods quantize point clouds into regular voxels in 3D space with a predefined resolution, and then apply volumetric convolution. More recently, some works explore new designs of local aggregation operators on point clouds to process point sets efficiently and reduce the loss of details~\cite{qi2017pointnet,qi2017PointNetplusplus,wang2019dynamic}. PointNet~\cite{qi2017pointnet} is a pioneering deep network architecture that directly processes point clouds for classification and semantic segmentation; it proposes a shared multi-layer perception (MLP), followed by a max-pooling layer, to approximate continuous set functions to deal with unordered point sets. %PointNet treats each point independently and ignores the geometric relationships among points, and thus only local features are extracted. 
PointNet++~\cite{qi2017PointNetplusplus} further proposes a hierarchical aggregation of point features to extract global features. In later works, DGCNN~\cite{wang2019dynamic} proposes an effective EdgeConv operator that encodes the point relationships as edge features to better capture local geometric features of point clouds while still maintaining permutation invariance. In this paper, we leverage a simplified PointNet architecture to extract local features from a point cloud before feeding it to an attention-based LSTM for point cloud sampling.

%and generates local subsets of points by sampling around key points. The features of those subsets are then grouped into sets for further feature extraction

\vspace{-5pt}
\paragraph{Point Cloud Sampling}\vspace{-5pt}
Random sampling (RS) selects a set of points randomly from a point cloud and has the smallest computation overhead. But this method is sensitive to density imbalance issue~\cite{lang2020SampleNet}. Farthest point sampling (FPS)~\cite{eldar1997FPS, moenning2003fast} has been widely used as a pooling operator in point cloud processing. It starts from a randomly selected point in the set and iteratively selects the next point from the point cloud that is the farthest from the selected points, such that the sampled points can achieve a maximal coverage of the input point cloud. %Voxel-based sampling methods~\cite{wu2015modelnet, qi2016volumetric} quantize point clouds into regular voxels in 3D space with a predefined resolution, and transfer point cloud to occupancy grid and apply volumetric convolution. However, voxel-based methods suffer from high computational cost since the network size and computational complexity grow quickly with spatial resolution.
% Recently, ~\cite{Xu_2020_CVPR} proposed a coverage-aware grid query (CAGQ) module for seeking optimal coverage over the original point cloud. The samplers are non-parametric and sensitive to outlier points, which means they may not be robust enough to handle real-world data. To overcome this drawback, [43] proposes a local adaptive shifting targeting on sampled points to improve the robustness and reduce the sensitivity to noisy points.  
% The learned task-oriented samplers have demonstrated superior performance over traditional task-agnostic sampling method, such as RS and FPS, in various applications.
Recently, S-NET~\cite{dovrat2019learning_to_sample} and SampleNet~\cite{lang2020SampleNet} have demonstrated that better sampling strategies can be learnt by a sampling network. These methods aim to generate a small set of samples that optimize the performance of a downstream task. Moreover, the generated 3D coordinates can be pushed towards a subset of original points to minimize the training loss if a matched point set is desired. Both methods treat the sampling process as a generation task and produce all the points in one shot, which does not pay sufficient attention to sample dependency, and leads to suboptimal performances. Our APSNet is a combination of FPS and task-oriented sampling in the sense that points are sampled sequentially with a task-oriented objective.

\vspace{-5pt}
\paragraph{Knowledge Distillation}\vspace{-5pt}
As one of the popular model compression techniques, knowledge distillation~\cite{kd14} is inspired by knowledge transfer from teachers to students. Its key strategy is to orientate compact student models to approximate over-parameterized teacher models such that student models can achieve the performances that are close to (sometimes even higher than) those of teachers'. %In knowledge distillation, soft targets from teacher network's predictions are used to train student networks since the soft targets are more informative than the one hot target labels.
Different from traditional knowledge distillation, which forces student networks to approximate the soft prediction of pre-trained teacher networks, self distillation~\cite{beownteacher} distills knowledge within a network itself from its own soft predictions. Our APSNet can be trained both in supervised learning and self-supervised learning via knowledge distillation. In the former case, labeled training data are required to train APSNet, while in the latter case the soft predictions of task network can be used to train APSNet such that the sampled point clouds from APSNet can achieve similar predictions as the original point clouds. 

% The networks are firstly divided into several sections. Then the knowledge in the deeper portion of the networks is squeezed into the shallow ones. It works like gradient injection or joint training in our case to help the gradient flow backwards more efficiently.

\vspace{-5pt}
\section{The Proposed Method}\vspace{-5pt}

The overview of our proposed APSNet is depicted in Fig.~\ref{fig:overview}, which contains two main components: (a) A simplified PointNet for feature extraction, (b) An LSTM with attention mechanism for sequential point sampling. In this section, we first describe the details of these components and then discuss different approaches to train APSNet.

\vspace{-5pt}
\paragraph{Notation and Problem Statement}\label{sec:notation}

Let $\bs{P} = \{\bs{p}_i\in\mathbb{R}^3\}_{i = 1}^n$ denote a point cloud that contains $n$ points, with $\bs{p}_i=[x_i,y_i,z_i]$ representing the 3D coordinates of point $i$. We consider two types of point cloud samples: (1) $\bs{Q}^* = \{\bs{q}_i^*\in \mathbb{R}^3\}_{i=1}^m$ denotes a \textbf{\emph{sampled}} point cloud of $m$ points that is a subset of $\bs{P}$ with $m<n$, i.e., $\bs{Q}^* \subset \bs{P}$. (2) $\bs{Q} = \{\bs{q}_i\in \mathbb{R}^3\}_{i=1}^m$ denotes a \textbf{\emph{generated}} point cloud of $m$ points that \emph{may not} be a subset of $\bs{P}$. Typically, we can convert $\bs{Q}$ to $\bs{Q}^*$ by a \emph{matching} process, i.e., match each point in $\bs{Q}$ to its nearest point in $\bs{P}$, and then replace the duplicated points\footnote{Multiple points in $\bs{Q}$ can be mapped to the same point in $\bs{P}$.} in resulting $\bs{Q}^*$ by FPS. Without loss of generality, in the following we present our algorithm in terms of $\bs{Q}$ since $\bs{Q}$ is more general than $\bs{Q}^*$ and can be converted to $\bs{Q}^*$ by matching. Moreover, let $f_{\bs{\theta}}(\cdot): \bs{P}\to\bs{Q}$ denote APSNet with the parameters $\bs{\theta}$.

As discussed in the introduction, we are interested in task-oriented sampling that yields a small set of points $\bs{Q}$ to optimize a downstream task represented by a well-trained deployed task network $T$, where $T$ can be a model for 3D point cloud classification, reconstruction or registration, etc. Given $\bs{P}$, the goal of APSNet is to generate a point cloud $\bs{Q}=f_{\bs{\theta}}(\bs{P})$ that maximizes the predictive performance of task network $T$. Specifically, the parameters of APSNet, $\bs{\theta}$, is optimized by minimizing a task loss and a sampling loss jointly as
\begin{equation} 
\min_{\bs{\theta}} \ell_{task}(T(\bs{Q}), y) + \lambda L_{sample}(\bs{Q},\bs{P}),
\end{equation}
whose details are to be discussed in Sec.~\ref{sec:objective}.

\vspace{-5pt}
\subsection{Attention-based Point Cloud Sampling}
Existing task-oriented sampling methods, such as S-NET~\cite{dovrat2019learning_to_sample} and SampleNet~\cite{lang2020SampleNet}, formulate the sampling process as a point cloud generation problem from a global feature vector, and generate all $m$ points in one shot. We argue that the sampling process is naturally a sequential generation process, in which points to be sampled next should depend on the points that have already been sampled. We therefore propose APSNet, an attention-based LSTM for sequential point sampling in order to capture the relationship among points. 

\begin{figure}[t]
\centerline{
\includegraphics[width = 0.65\linewidth]{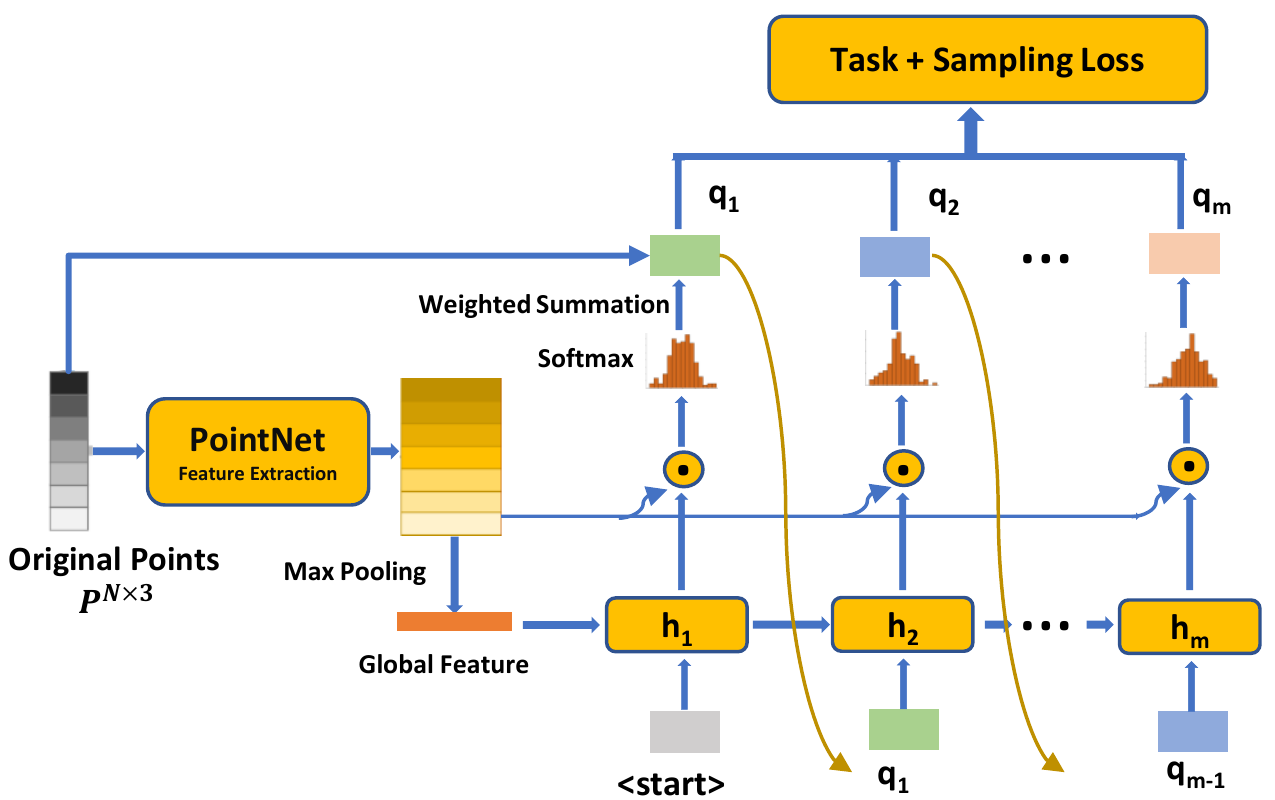}}\vspace{-10pt}
\caption{APSNet considers point cloud sampling as a sequential generation process with a task-oriented objective, and uses an LSTM with attention mechanism for sampling.}
\label{fig:lstm}
%\vspace{-5pt}
\end{figure}

The overall architecture of APSNet is depicted in Fig.~\ref{fig:lstm}. First, APSNet takes the original point cloud $\bs{P}$ as input and samples from $\bs{P}$ via an LSTM with attention mechanism to produce a small point cloud $\bs{Q}$ of $m$ points, each point of which is a soft point generated by projecting $\bs{P}$ on a set of attention coefficients from the LSTM. Finally, the output of APSNet, $\bs{Q}$, is fed to a well-trained deployed task network $T$ for prediction and task loss evaluation\footnote{The parameters of $T$ is frozen during the training of APSNet.}.

The first step is to extract a feature representation for each point in $\bs{P}$. APSNet follows the architecture of PointNet~\cite{qi2017pointnet}, a basic feature extraction backbone on 3D point clouds, to extract point-wise local features. Specifically, a set of $1 \times 1$ convolution layers are applied to the original point cloud $\bs{P}$ and produce a set of $d$-dimensional point-wise feature vectors, denoted by $\bs{X}=[\bs{x}_1,\bs{x}_2,\cdots,\bs{x}_n]^\textsf{T}\in\mathbb{R}^{n\times d}$. Then, a symmetric feature-wise max pooling operator is applied to $\bs{X}$ and produce a global feature vector $\bs{g}\in\mathbb{R}^d$, which is then fed to an LSTM as the initial state for sequential point generation.

The sequential point generation process is similar to the attention-based sequential model for image captioning~\cite{Xu2015}. Given initial state $\bs{g}$ and \texttt{<start>} as inputs, the LSTM updates its hidden state to $\bs{h}_t\in\mathbb{R}^d$ at each time step $t=1,2,\cdots,m$. The hidden state $\bs{h}_t$ encodes the history of all the sampled points and is indicative for APSNet to identify the next most informative point of $\bs{P}$ to sample. To achieve this, a set of attention scores is calculated as the dot product of the hidden state $\bs{h}_t$ and point-wise feature vector $\bs{x}_i$ for $i=1,2,\cdots, n$, followed by a softmax for normalization:\vspace{-5pt}
\begin{align}
  s_{it}=\bs{x}_i \cdot \bs{h}_t, \;\;\;\;\;\;\;\; a_{it}=\frac{\exp(s_{it})}{\sum_i\exp(s_{it})}.\vspace{-5pt}
\end{align}
The attention coefficients $a_{it}$ indicates the importance of point $i$ at the sampling step $t$, from which sampled point $\bs{q}_t\in\mathbb{R}^3$ can be generated as a weighted sum of all the points in $\bs{P}$:\vspace{-5pt}
\begin{align}
  \bs{q}_t&=\sum_{i}a_{it}\cdot \bs{p}_i. \vspace{-10pt}
\end{align}
The generated point $\bs{q}_t$ is then fed to the LSTM as input for the next time step to generate the next point until all $m$ points are generated. During sequential generation process, the attention mechanism enables the model to attend to all the points in $\bs{P}$ and identify the most informative ``point" to sample.

\vspace{-5pt}
\subsection{Training of APSNet}\label{sec:objective}\vspace{-5pt}
%As discussed in the introduction, 
APSNet is a task-oriented sampling network, which can be trained to optimize its performance on the downstream tasks of interest. In this section, we discuss the objective functions that can be used to train APSNet. Depending on the availability of labeled training data and deployment requirements, we consider three different approaches to train APSNet: (1) supervised training, (2) self-supervised training, and (3) joint training.

\vspace{-5pt}
\subsubsection{Training with or without Ground Truth Labels}\vspace{-5pt}
We consider two training scenarios: (a) both task network $T$ and a labeled training set $\{\bs{P}_j, y_j\}_{j=1}^N$ are available; (b) only task network $T$ and some input point clouds $\{\bs{P}_j\}_{j=1}^N$ are available, but no labels is provided. The latter case corresponds to the situation where original labeled training data of $T$ is no longer available for the development of APSNet.

\vspace{-10pt}
\paragraph{Supervised Training}
When a labeled training set $\{\bs{P}_j,y_j\}_{j=1}^N$ is available, we can train APSNet in a supervised learning paradigm. Similar to S-NET~\cite{dovrat2019learning_to_sample} and SampleNet~\cite{lang2020SampleNet}, two types of losses are exploited to train APSNet, i.e., the task loss $\ell_{task}$ and the sampling loss $L_{sample}$. Specifically, the task loss measures the quality of predictions based on the sampled point cloud $\bs{Q}$:\vspace{-5pt}
\begin{equation}
    \ell_{task}(T(\bs{Q}), y),
\end{equation}
where $y$ is the ground-truth label of $\bs{P}$. For different downstream tasks, $y$ can be the class label or the original point cloud $\bs{P}$ when the task is for classification or reconstruction, respectively. Accordingly, the corresponding task loss $\ell_{task}()$ is defined differently, e.g., the cross-entropy loss for classification or the Chamfer distance for reconstruction~\cite{achlioptas2018learning}.

The sampling loss $L_{sample}$ encourages the sampled points in $\bs{Q}$ to be close to those of $\bs{P}$ and also have a maximal coverage w.r.t. the original point cloud $\bs{P}$. We found that this sampling loss provides an important prior knowledge for sampling, and is critical for APSNet to achieve a good performance. Specifically, given two point sets $\bs{S}_1$ and $\bs{S}_2$, denoting average nearest neighbor loss as:%\vspace{-5pt}
\begin{align} \label{eq:loss_a}
L_a(\bs{S}_1,\bs{S}_2) = \frac{1}{|\bs{S}_1|}\sum_{s_1 \in \bs{S}_1}{\min_{s_2 \in \bs{S}_2}||s_1-s_2||_2^2},
\end{align}
and maximal nearest neighbor loss as:%\vspace{-5pt}
\begin{align} \label{eq:loss_m}
L_m(\bs{S}_1,\bs{S}_2) = \max_{s_1 \in \bs{S}_1}{\min_{s_2 \in \bs{S}_2}||s_1-s_2||_2^2},
\end{align}
the sampling loss is then given by:\vspace{-5pt}
\begin{align} \label{eq:loss_simplify}
L_{sample}(\bs{Q},\bs{P}) = &L_a(\bs{Q},\bs{P}) + \beta L_m(\bs{Q},\bs{P}) + (\gamma + \delta|\bs{Q}|) L_a(\bs{P},\bs{Q}),
\end{align}
where $\beta$, $\gamma$ and $\delta$ are the hyperparameters that balance the contributions from different loss components.
% In our experiments, we set ...

Putting all the components together, the total loss of APSNet is given by:%\vspace{-5pt}
\begin{align} \label{eq:loss_total_sample}
L_{total} = \ell_{task}(T(\bs{Q}), y) + \lambda L_{sample}(\bs{Q},\bs{P}),\vspace{-10pt}
\end{align}
where $\lambda$ is a hyperparameter that balances the contribution between the task loss and the sampling loss.

\paragraph{Self-supervised Training with Knowledge Distillation}
In some practical scenarios, we may only have task network $T$ and some input point clouds $\{\bs{P}_j\}_{j=1}^N$ at our disposal. This is the situation where a deployed task network $T$ is available, but the original labeled training data of $T$ is no longer available for the development of APSNet. In this case, we propose to train APSNet via self-supervised learning based on the idea of knowledge distillation~\cite{kd14,beownteacher}. In knowledge distillation, we can transfer the knowledge from a teacher network to a student network such that the student network can yield a similar prediction as the teacher network while being much more efficient. Inspired by knowledge distillation, we treat the task network $T$ as the teacher model, and APSNet as the student model and use the soft predictions of $T$ as the targets to train APSNet. Specifically, the task loss for self-supervised training of APSNet is updated to %\vspace{-10pt}
\begin{align} \label{eq:kd}
\ell_{task}(T(\bs{Q}), \tilde{y}), \text{ with } \tilde{y}=T(\bs{P}),
\end{align}
where $\tilde{y}$ is the soft prediction of $T$ given the original point cloud $\bs{P}$, and the loss function $\ell_{task}$ is defined differently for different downstream tasks. The goal of the new loss function is to generate a sampled point cloud $\bs{Q}$ that can yield a similar prediction as the original $\bs{P}$.

% , e.g., the KL divergence for classification or the Chamfer Distance loss for reconstruction. 

Similar idea is also explored in~\cite{l2x18}, where mutual information between the predictions of backbone network from sparsified input and original input are maximized for model interpretation, while here we sparsify point clouds.

%According to~\cite{l2x18}, we can maximize the mutual information between the output of backbone network from sampled points and the whole point set. We use the prediction from backbone network as target to train the sampler. This is also similar to knowledge distillation. Backbone network is used as the teacher model to compress the input point set. According to~\cite{kd14,beownteacher}, the soft target used to train the student model contains more useful information for training smaller model and can even improve the performance.

% From the experiments, we find that sequential generation works better to get the first small group of points. In order to reduce the generation time, we can generate 128 points and use FPS to complete the points. This method works very well in all our experiments.

\vspace{-5pt}
\subsubsection{Joint Training}\vspace{-5pt}
APSNet described above is trained for a specified sample size $m$. Given the autoregressive model of our method, APSNet can generate arbitrary length of samples from a single model. This entails an effective joint training of APSNet with multiple different sample sizes, resulting in a single compact model to generate arbitrary sized point clouds with prominent performance. Specifically, we can train one APSNet with different sample sizes by:%\vspace{-5pt}
\begin{equation} \label{eq:loss_joint}
L_{joint}\!\! =\!\!\!\!\sum_{c \in C_s}\!\!\Big(\ell_{task}(T(\bs{Q}_c),y) + \lambda L_{sample}(\bs{Q}_c,\bs{P})\Big),%\vspace{-15pt}
\end{equation}
where $C_s$ is a set of sample sizes of interest. In our experiments, we set $C_s = \{2^l\}_{l=3}^{7}$.

S-NET~\cite{dovrat2019learning_to_sample} and SampleNet~\cite{lang2020SampleNet} propose a progressive training to train a sampling network to generate different sized point clouds. However, their model sizes grow linearly as the sample size $m$ increases. In contrast, due to the autoregressive model of APSNet, we can train one single compact model (with a fixed number of parameters) to generate arbitrary sized point clouds without incurring a linear increase of model parameters. This entails a better parameter reusing or sharing for APSNet, and leads to improved sample efficiency as compared to S-NET and SampleNet.

%SampleNet prefers SampleNet-M for a better and consistent performance, while APSNet prefers APSNet-G as it achieves higher accuracy without an expensive matching. 

%\vspace{-5pt}
\section{Experiments}%\vspace{-5pt}
We demonstrate the performance of APSNet on three different applications of 3D point clouds for classification, reconstruction, and registration. For the purpose of comparison, random sampling (RS), farthest point sampling (FPS) and SampleNet~\cite{lang2020SampleNet} are used as baselines, where SampleNet is the state-of-the-art task-oriented sampling method. We consider two variants of APSNet: (1) \textit{APSNet}, and (2) \textit{APSNet-KD}, while the former refers to the supervised training of APSNet and the latter refers to the self-supervised training of APSNet with knowledge distillation. A trained APSNet generates point cloud $\bs{Q}$ that isn't a subset of original input point cloud $\bs{P}$, but the generated $\bs{Q}$ can be converted to $\bs{Q}^*$ by a matching process as discussed in Sec.~3.1. %~\ref{sec:notation}. 
Therefore, we further distinguish them as \textit{APSNet-G} and \textit{APSNet-M}, respectively. SampleNet~\cite{lang2020SampleNet} also generates point cloud $\bs{Q}$, which is converted to $\bs{Q}^*$ by the matching process. Similarly, we denote them as \textit{SampleNet-G} and \textit{SampleNet-M}, respectively. In our experiments, we compare the performances of all these variants. However, we would like to emphasize that the default SampleNet is \textit{SampleNet-M}, while the default APSNet is \textit{APSNet-G} since APSNet-G yields the best predictive performance without an expensive matching process as we will demonstrate in the experiments.

Since our APSNet is implemented in PyTorch, we convert the official TensorFlow implementation of SampleNet\footnote{\url{https://github.com/itailang/SampleNet}} to PyTorch for a fair comparison. We found that our PyTorch implementation achieves better performances than the official TensorFlow version in most of our experiments. Details of experimental settings and implementation are relegated to supplementary material. All our experiments are performed on Nvidia RTX GPUs. Our source code can be found at \url{https://github.com/Yangyeeee/APSNet}.

% \begin{table}[h] \label{hyper_parameters}
% \vspace{-5px}
% \begin{center}
% \begin{tabular}{ c c c c c }
% \hline
%   &  Classification   &   Registration    &     Reconstruction    \\
% \hline
% % $k$              &   $7$             &   $8$         &       $16$            \\
% $\lambda$         &   $30$            &   $0.01$      &       $0.01$          \\
% $\beta$          &   $1$             &   $1$         &       $1$             \\
% $\gamma$         &   $1$             &   $1$         &       $0$             \\
% $\delta$         &   $0$             &   $0$         &       $0$          \\
% $bs$      &   $128$            &   $128$        &       $128$            \\
% $lr$    &   $0.01$          &   $0.001$     &       $0.0005$        \\
% \hline
% \end{tabular}
% \end{center}
% \caption{Hyperparameters for the training of APSNet on three point cloud applications. $bs$: batch size, $lr$: learning rate.}
% \label{table:hyper_parameters}
% \vspace{-10px}
% \end{table}

%\vspace{-5pt}
\subsection{3D Point Cloud Classification}%\vspace{-5pt}
We use the point clouds of 1024 points that were uniformly sampled from the ModelNet40 dataset~\cite{wu2015modelnet} to train PointNet~\cite{qi2017pointnet} (the task network $T$). The official train-test split is used for the training and evaluation, and the instance-wise accuracy is used as the evaluation metric for performance comparison. The vanilla task network achieves an accuracy of 90.1\% with all the 1024 points. We execute different sampling methods with a variety of sample sizes and report their performances for comparison.

%Vanilla task network can achieve an accuracy of 90.1\% with all the 1024 points. 

\vspace{-0pt}
\begin{table*}[h!]
\centering
%\begin{adjustbox}{width=\columnwidth,center}
\caption{Classification accuracies of five sampling methods with different sample sizes $m$ on ModelNet40. M* denotes the official results from SampleNet~\cite{lang2020SampleNet}.}
\scalebox{0.85}{
\begin{tabular}{c|c|c|c|cc|ccc|cc|cc}\Xhline{5\arrayrulewidth}
%\backslashbox{Methods}{Metrics}
   &\makebox[2em]{RS}&\makebox[2em]{FPS} &\makebox[2em]{DaNet} & \multicolumn{2}{c|}{\makebox[2em]{MOPS-Net~\cite{qian2020mops}}} &  \multicolumn{3}{c|}{\makebox[2 em]{SampleNet}}  & \multicolumn{2}{c|}{\makebox[2em]{APSNet}} &  \multicolumn{2}{c}{\makebox[2em]{APSNet-KD }}  \\
      $m$ & & & \makebox[2em]{~\cite{9551070}} &\makebox[2em]{G}&\makebox[2em]{M} &\makebox[2em]{G}&\makebox[2em]{M}& \makebox[2em]{M*}&\makebox[2em]{G}&\makebox[2em]{M}   &\makebox[2em]{G}&\makebox[2em]{M} \\\Xhline{2\arrayrulewidth}
8&       8.26 &    23.29& -& -& -&  78.36&    73.31& 28.7&  \textbf{81.42}&  74.12&    80.22&   73.81\\
16&     25.11&     54.19& -& \textbf{84.7}& 51.2&  80.60&    79.68& 55.5&  83.89&  82.25&    83.82&   82.02\\
32&     55.19&     77.32& 85.1& 86.1& 77.6&  80.32&    82.97& 74.4&  88.15&  86.97&    \textbf{88.76}&   84.95 \\
64&     78.26&     87.22& 86.8& 87.1& 81.0&  79.36&    84.01& 79.0&  88.38&  87.58&    \textbf{88.66}&   87.54  \\
128&    85.95&     88.76& 86.8& 87.2& 85.0&  85.52&    87.17& 79.7&  89.22&  \textbf{89.38}&    87.83&   88.01   \\
256&    88.80&     89.30& 87.2& 87.4& 86.7&  87.43&    89.58& 83.4&  89.54&  \textbf{89.86}&    88.02&    88.21 \\
512&    89.66&     89.87& -& 88.3& 88.3&  88.01&    90.18& 88.2&  89.78&  \textbf{90.18}&    88.69&    88.56\\
\Xhline{5\arrayrulewidth}
\end{tabular}}
%\end{adjustbox}%
\label{table:classification}
\vspace{-0pt}
\end{table*}

Table~\ref{table:classification} reports the classification accuracies of all the five sampling methods. To validate our PyTorch implementation of SampleNet, we also include the official SampleNet results as reported in~\cite{lang2020SampleNet}. It can be observed that our PyTorch implementation outperforms the official TensorFlow code consistently; in particular, when sample size $m=8$, our implementation has a gain of nearly 45\% over the official code. Therefore, for a fair comparison, we compare APSNet mainly with our improved SampleNet.

A few notable observations can be made from Table~\ref{table:classification}. (1) As sample size $m$ increases, all the sampling methods have improved accuracies. The performances of task-oriented samplers, e.g., SampleNet and APSNet, outperform those of task-agnostic samplers, e.g., random sampling and FPS, consistently. However, the gains are getting smaller as sample size $m$ increases; when $m=512$, all sampling methods achieves a comparable accuracy that is close to the best accuracy (90.1\%) achieved with all the 1024 points. (2) In general, SampleNet-M achieves a better performance than SampleNet-G. When sample size $m$ increases, the gain is more pronounced. (3) In contrast, APSNet-G achieves a better performance than APSNet-M. When sample size $m$ is small, the gain is large, while as sample size $m$ increases, both variants of APSNet have very similar performances. This is likely because the downstream task networks are trained with original points $P$, and the matched $Q^*$ from APSNet-M can fit better to the downstream task networks. The gains are getting smaller because when sampling ratio becomes larger the performance is approaching to the upper bound which uses all the points. (4) Comparing APSNet-G with SampleNet-M (the best defaults for both algorithms), APSNet outperforms SampleNet consistently; especially when $m\leq128$, we observe a 2\% to 8\% accuracy gain, demonstrating the effectiveness of APSNet. (5) APSNet-KD achieves a very impressive result without utilizing labeled point clouds for training; its performance is almost on-par with APSNet that is trained with labeled point clouds.

% For different sampling size m, we can achieve at most  5\% improvement when m = 32. We found that when sampling number is small, using the original points will not achieve the optimal performance, it will have 1\% or 2\% performance drop comparing with using soft points for APSNet. The SampleNet-G performance is not good and the accuracy is not consistently increased when more points are used. 

\vspace{-10pt}
\paragraph{Discussion}
The above experiments show that SampleNet-M outperforms SampleNet-G consistently, while the opposite is observed for APSNet. This can be explained by the limitations of SampleNet as we indicated in the introduction. As sample size $m$ increases, SampleNet has a higher probability of generating similar (redundant) points due to the one-shot generation of $m$ samples. Since these redundant points cannot improve the classification accuracy effectively, the matching process (which replaces the redundant points with the FPS samples) becomes critical for SampleNet-M to improve its performance, leading to improved performances over SampleNet-G. On the other hand, APSNet-G generates the next sample depending heavily on previously sampled points, and therefore is able to capture the relationship among points and generate more informative samples, yielding a better performance without an expensive matching process.

%This indicates that the one-shot generation process of SampleNet can not capture the relationship among points well and sampled redundant points undermine the performance of SampleNet.

%\paragraph{Self-supervised training} For APSNet-KD, we use the predicted soft targets for the whole 1024 points as the labels to train the model. The performance are similar to supervised training. This may be explained by that the task network has the ability to identify the importance of different extracted feature. By minimizing the difference of feature distribution from 1024 points and the sampled points, we are trying to maximize the mutual information between the original point cloud and sampled points. By doing so, we can achieve reasonable results without ground truth label. The results are showing in the last column in Table~\ref{table:classification}

%\begin{wrapfigure}{R}{0.3\textwidth}%\vspace{-80pt}
%\centerline{
%\includegraphics[width =4.5cm]{pic/joint.png}}\vspace{-10pt}
%\caption{Evolution of classification accuracy as a function of sample size $m$ for different sampling methods.} 
%\label{fig:classification}
%\vspace{-10pt}
%\end{wrapfigure}

\begin{figure}
\centerline{
\includegraphics[width =6.0cm]{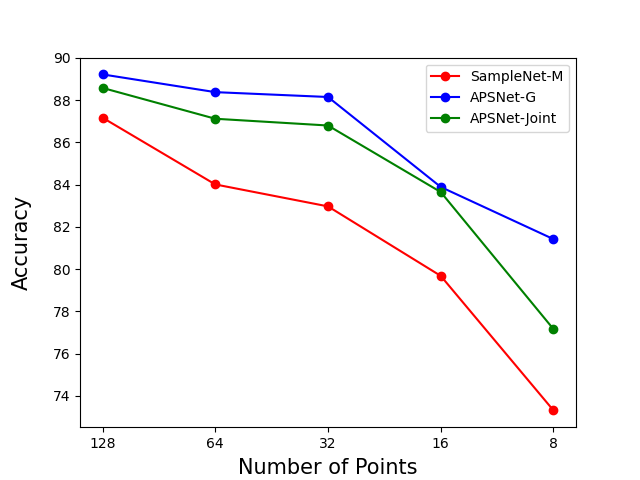}}\vspace{-10pt}
\caption{Evolution of classification accuracy as a function of sample size $m$ for different sampling methods. APSNet-Joint uses a single model to generate different number of samples, while SampleNet and APSNet use separately trained models to generate each specific number of samples.}
\label{fig:classification}
%\vspace{-10pt}
\end{figure}

\vspace{-5pt}
\paragraph{Joint Training} Next, we investigate the joint training of APSNet, and compare it with separated training of ASPNet and SampleNet for each sample size $m$. For the joint training of APSNet, we train a single compact model of APSNet with $C_s=\{8, 16, 32, 64, 128\}$ by optimizing the joint loss (\ref{eq:loss_joint}). In contrast, in separated training of SampleNet or APSNet, a separated model is trained for each sample size $m\in C_s$ and its performance is evaluated for the specific $m$ it was trained with.

Fig.~\ref{fig:classification} shows the performance comparison between joint training of APSNet and separated training. It can be observed a single model trained by APSNet-Joint can generate arbitrary length of samples with competitive performances. Indeed, the performance of APSNet-Joint is lower than the separately trained APSNet, but it still consistently outperforms separately trained SampleNet. 

%The progressive training of in ~\cite{lang2020SampleNet} achieve similar results as their independent model. We use there independent model results to compare with our joint training results. We still have a big gain, which demonstrate the advantage of using LSTM for sampling to generate any output number of points without introducing more parameters.

\vspace{-5pt}
\subsection{Additional Experiments}\vspace{-5pt}
Due to page constraints, additional experiments of (1) point cloud sampling for reconstruction,  (2) registration, (3) ablation study of the sampling loss~(\ref{eq:loss_simplify}), (4) inference speed comparison, and (5) visualization of attention coefficients are relegated to supplementary material.

\vspace{-5pt}
\section{Conclusion}\vspace{-5pt}
This paper introduces APSNet, an attention-based sampling network for point cloud sampling. Compared to S-Net and SampleNet, which formulate the sampling process as an one-shot generation task with MLPs, APSNet employs a sequential autoregressive generation with a novel LSTM-based sequential model for sampling. Depending on the availability of labeled training data, APSNet can be trained in supervised learning or self-supervised learning via knowledge distillation. We also present a joint training of APSNet, yielding a single compact model that can generate arbitrary length of samples with prominent performances. Extensive experiments demonstrate the superior performance of APSNet over the state-of-the-arts both in terms of sample quality and inference speed, which make APSNet widely applicable in many practical application scenarios. 

\section{Acknowledgements }\vspace{-5pt}
We would like to thank the anonymous reviewers for their comments and suggestions, which helped improve the quality of this paper. We would also gratefully acknowledge the support of Cisco Systems, Inc. for its university research fund to this research.

\bibliography{egbib}

\newpage
\setcounter{section}{6}
\section{Appendix}~\label{sec:app}

\vspace{-30pt}
\subsection{Experimental Settings}~\label{app:exp}
\vspace{-25pt}

\paragraph{Task Networks} Similar to SampleNet~\cite{lang2020SampleNet}, we adopt PointNet for classification~\cite{qi2017pointnet}, Point Cloud Autoencoder (PCAE) for reconstruction~\cite{achlioptas2018learning}, and PCRNet for registration~\cite{sarode2019pcrnet}. For the classification and reconstruction tasks, PointNet and PCAE are trained with the same settings as reported by their original papers. For the registration task, Sarode et al.~\cite{sarode2019pcrnet} trained the PCRNet with the Chamfer distance between template point cloud and registered point cloud; we follow SampleNet and add a regression loss (besides the Chamfer distance) to train the PCRNet. These pre-trained networks are treated as the task networks for their specific applications, whose parameters are fixed during the training of APSNet.

%between the estimated transformation and the ground truth transformation

\vspace{-10pt}
\paragraph{APSNet} The feature extract component of APSNet follows the design of PointNet~\cite{qi2017pointnet}. It contains a sequence of $1\times 1$ convolution layers, followed by a symmetric global pooling layer to generate a global feature vector, which is then used as the initial state of LSTM for sampling. Each convolution layer includes a batch normalization layer~\cite{ioffe2015batch} and a ReLU activation function. A 2-layer LSTM~\cite{lstm} with 128 recurrent units in each layer is used to generate samples autoregressively. %From the experiments, we find that sequential generation works better to get the first small group of points. In order to reduce the generation time, we generate 128 points and use FPS to complete the points. This method works very well in all our experiments. 

We consider two variants of APSNet: (1) \textit{APSNet}, and (2) \textit{APSNet-KD}, while the former refers to the supervised training of APSNet and the latter refers to the self-supervised training of APSNet with knowledge distillation. A trained APSNet generates point cloud $\bs{Q}$ that isn't a subset of original input point cloud $\bs{P}$, but the generated $\bs{Q}$ can be converted to $\bs{Q}^*$ by a matching process as discussed in Sec.~3.1. %~\ref{sec:notation}. 
Therefore, we further distinguish them as \textit{APSNet-G} and \textit{APSNet-M}, respectively. 

SampleNet~\cite{lang2020SampleNet} also generates point cloud $\bs{Q}$, which is converted to $\bs{Q}^*$ by the matching process. Similarly, we denote them as \textit{SampleNet-G} and \textit{SampleNet-M}, respectively. In our experiments, we compare the performances of all these variants. However, we would like to emphasize that the default SampleNet is \textit{SampleNet-M}, while the default APSNet is \textit{APSNet-G} since APSNet-G yields the best predictive performance without an expensive matching process as we will demonstrate in the experiments.

%For the rest part, \textbf{SampleNet-M} refers to the original SampleNet, which will do matching process during inference for performance improvement. \textbf{SampleNet-G} means SampleNet using generated soft points. \textbf{APSNet-G} is the model we can achieve the best performance and \textbf{APSNet-M} will do matching and completing for fair comparison with SampleNet.\textbf{APSNet-KD} refers to APSNet  self-supervised training APSNet with knowledge distillation
% In order to improve the efficiency to generate a large number of samples, we generate at most 128 points by APSNet and sample the rest of points by FPS.

%The initial input point is randomly initialized. Table~\ref{hyper_parameters} lists the hyperparameters for the optimization of APSNet.

\vspace{-5pt}
\paragraph{Implementation} We tune the performance of APSNet based on the hyperparameters provided by SampleNet~\cite{lang2020SampleNet}, and set $\beta$ = 1, $\gamma$ = 1 and $\delta$ = 0. We use the Adam optimizer~\cite{kingma2014adam} with the batch size of 128 for all the experiments. Learning rate is set to (0.01, 0.001, 0.0005), and $\lambda$ of the total loss (9)  %(\ref{eq:loss_total_sample}) 
is set (30, 0.01, 0.01) for classification, registration, and reconstruction tasks, respectively. Each experiment is trained for 400 epochs with a learning rate decay of $0.7$ at every $20$ epochs. 

Since our code is PyTorch-based, we convert the official
TensorFlow code of SampleNet\footnote{\url{https://github.com/itailang/SampleNet}} to PyTorch for a fair comparison. We found that our PyTorch implementation achieves better performances than the official TensorFlow version in most of our experiments. For reproducibility, our source code is also provided as a part of the supplementary material. All our experiments are performed on Nvidia RTX GPUs.

%\vspace{-5pt}
\subsection{Reconstruction}\vspace{-5pt}

The reconstruction task is evaluated with point clouds of 2048 points, sampled from the ShapeNet Core55 dataset~\cite{chang2015shapenet}. We choose the four shape classes that have the largest number of examples: Table, Car, Chair, and Airplane. Each class is split to a 85\%, 5\%, 10\% partition for training, validation and test. The task network, in this case, is the Point Cloud Autoencoder (PCAE) for reconstruction~\cite{achlioptas2018learning}. We evaluate the reconstruction performance with the normalized reconstruction error (NRE):\vspace{-5pt}
\begin{equation}
    \textsf{NRE}_\textsf{CD}(\bs{Q}, \bs{P})=\frac{\textsf{CD}(\bs{P}, T(\bs{Q}))}{\textsf{CD}(\bs{P}, T(\bs{P}))},
    \label{eqn:nre}\vspace{-5pt}
\end{equation}
where \textsf{CD} is the Chamfer distance~\cite{achlioptas2018learning} between two point clouds. Apparently, the values of $\textsf{NRE}_{\textsf{CD}}$ are lower bounded by 1, and the smaller, the better.

\begin{table*}[t]
\centering
%\begin{adjustbox}{width=\columnwidth,center}
\caption{The normalized reconstruction errors of five sampling methods with different sample sizes $m$ on the ShapeNet Core55 dataset. M$^*$ denotes the original results from the SampleNet paper~\cite{lang2020SampleNet}. The lower, the better.}
\scalebox{0.85}{
\begin{tabular}{c|c|c|ccc|cc|cc}\Xhline{5\arrayrulewidth}
%\backslashbox{Methods}{Metrics}
   &\makebox[2.7em]{RS}&\makebox[2.7em]{FPS} &  \multicolumn{3}{c|}{\makebox[2.7 em]{SampleNet}}  & \multicolumn{2}{c|}{\makebox[2.7em]{APSNet}} &  \multicolumn{2}{c}{\makebox[2.7em]{APSNet-KD}}  \\
      $m$ & & &\makebox[2.7em]{G}&\makebox[2.7em]{M}& \makebox[2.7em]{M*}&\makebox[2.7em]{G}&\makebox[2.7em]{M}   &\makebox[2.7em]{G}&\makebox[2.7em]{M} \\\Xhline{2\arrayrulewidth}
8&    21.85& 12.79&   5.29 &   5.48& -&   \textbf{4.27}&  4.59&   4.69&  4.98   \\
16&   13.47& 7.25&    2.78 &   2.89& -&   \textbf{2.51}&  2.62&   2.57&  2.67   \\
32&   8.16&  3.84&    1.68 &   1.71& 2.32&   1.54&  1.59&  \textbf{1.47}&   1.52    \\
64&   4.54&  2.23&    1.32 &   1.27& 1.33&    \textbf{1.07}&  1.11&  1.12&   1.14     \\
% 128&   -   &   -   &      -   &   89.34&  17.67&  89.22&  13.4&   88.58&  89.38&  87.83&   8.29   \\
% 256&    -  &   -   &      -   &   89.82&  17.67&  89.54&  13.4&   88.02&  89.86&  88.02&   8.29  \\
% 512&    -  &   -   &      -   &   90.06&  17.67&  89.78&  13.4&   87.18&  90.18&  8.69&   8.29\\
\Xhline{5\arrayrulewidth}
\end{tabular}}
%\end{adjustbox}%
\label{table:reconstruction}
\vspace{-5pt}
\end{table*}

Table~\ref{table:reconstruction} reports the reconstruction results of all the five sampling methods considered. Similar to the results of classification, (1) SampleNet and APSNet outperform RS and FPS by a large margin. (2) SampleNet-M relies on the matching process to replace the redundant samples by FPS to improve its performance over SampleNet-G. (3) In contrast, APSNet-G outperforms APSNet-M consistently without the extra matching process. (4) APSNet-KD again achieves a very competitive result to APSNet. (5) Comparing APSNet-G and SampleNet-M (the best defaults for both algorithms), APSNet outperforms SampleNet consistently by a notable margin. 

\begin{figure*}[t]
\centerline{
\includegraphics[width = 12cm]{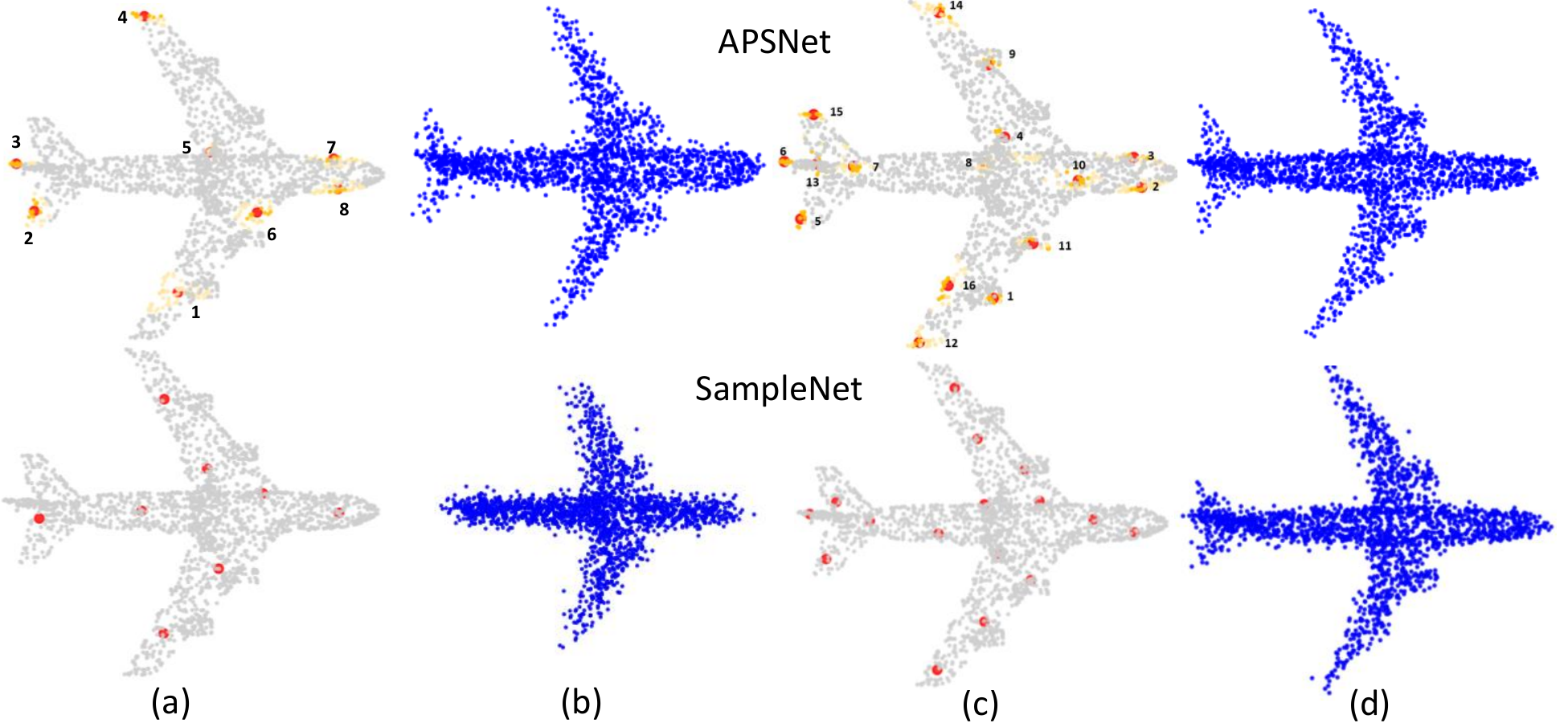}}\vspace{-10pt}
\caption{Visualization of sampled points and reconstructed point clouds by APSNet (1st row) and SampleNet (2nd row). The red dots are the sampled points; the highlighted yellow regions in APSNet results are points with high attention scores and the number specify the order of sampled points. (a) Sampled points when $m=8$; (b) Reconstruction when $m=8$, NRE(APSNet)=2.55, NRE(SampleNet)=5.20; (c) Sampled points when $m=16$; (d) Reconstruction when $m=16$, NRE(APSNet)=1.57, NRE(SampleNet)=2.34.}
%For this reconstruction task, SampleNet focuses more on the main body of airplane and samples some redundant symmetric points, while APSNet focuses more on the outline of airplane without losing details for reconstruction.
\label{fig:reconstruction}
\vspace{-10pt}
\end{figure*}

To investigate why APSNet outperforms SampleNet in the task of reconstruction, we visualize the sampled points and the reconstructed point clouds of both algorithms in Fig.~\ref{fig:reconstruction}. As can be seen, SampleNet focuses more on the main body of airplane and samples some uninformative and symmetric points for reconstruction. In contrast, APSNet focuses more on the outline of the airplane without losing details, which are critical for the reconstruction. This observation is more pronounced when sample size is small, such as $m=8$. As shown in Fig.~\ref{fig:reconstruction}(a) and (b), SampleNet fails to sample a point at the tail of the airplane such that the reconstructed point cloud cannot recover the tail. In comparison, APSNet samples two important points at the tail and ignores the symmetric one on the other side of the tail, and therefore is able to reconstruct the tail precisely. One the other hand, SampleNet samples two symmetric points on the wing, which are likely redundant information for the reconstruction. Overall, the sampled points from APSNet are more reasonable than those of SampleNet from human's perspective,

\paragraph{The effectiveness of different loss components}

The sampling loss~(7) encourages the sampled points in $\bs{Q}$ to be close to those of $\bs{P}$ and also have a maximal coverage w.r.t. the original point cloud $\bs{P}$. We found that this sampling loss provides an important prior knowledge for sampling, and is critical for APSNet to achieve a good performance. In addition, the sampling loss~(7) is more generic than the Chamfer distance since when $\beta$ = 0, $\gamma$ = 1 and $\delta$ = 0 it degenerates to the Chamfer distance. The limitation is that we now have more hyperparameters to tune. Table~\ref{tab:sample_loss} reports the ablation study of the sampling loss~(7) for APSNet-G on the reconstruction task. It shows that when $L_m(\bs{Q},\bs{P})$ and $L_a(\bs{P},\bs{Q})$ enabled (i.e., $\beta=1$ and $\gamma=1$), APSNet-G reaches the best results in almost all settings.

\begin{table}[h!]
\centering
\caption{Ablation study of the sampling loss~(7) for APSNet-G on the reconstruction task. * denotes the best results when $\beta$ = 1, $\gamma$ = 1 and $\delta$ = 0.}\label{tab:sample_loss}
\begin{tabular}{c|c|c|c}\Xhline{3\arrayrulewidth}
%\backslashbox{Methods}{Metrics}
  &\makebox[2.7em]{$\beta$ = 0}&\makebox[2.7em]{$\gamma$ = 0} &  \makebox[2.7em]{*} \\\Xhline{1\arrayrulewidth}
8&    4.63 &  4.54 &  4.27    \\
16&   3.07  & 3.44 &  2.51  \\
32&   1.67 & 1.49  &  1.54     \\
64&   1.13  & 1.28 &  1.07        \\
\Xhline{3\arrayrulewidth}
\end{tabular}
%\end{adjustbox}%
\end{table}
\vspace{-5pt}

\vspace{-5pt}
\paragraph{Inference time comparison} We further evaluate the inference times of different sampling methods in the task of reconstruction. The results are reported in Table~\ref{table:time}, where SampleNet-M and APSNet-G are the main algorithms to be compared since they are the best defaults. It can be observed that when sample size $m$ increases, the inference times of both SampleNet-M and APSNet-G increase, while SampleNet-G requires roughly a constant time for sampling. This is because SampleNet-G leverages an MLP generator to generate all $m$ samples in one shot; for the problem size considered, one GPU is able to utilize its on-board parallel resources to process different sample sizes in roughly the same time. However, as observed in the experiments above and also proposed by SampleNet~\cite{lang2020SampleNet}, SampletNet relies on the matching process to improve its performance, while matching is the most expensive operation in SampleNet, leading to a dramatic increase of inference-time for SampleNet-M. By contrast, due to the autoregressive model of our method, APSNet generates samples sequentially by an LSTM which results in a linear increase of inference time as $m$ increases. However, APSNet does not need an expensive matching process for its best performance. Therefore, besides the improved sample quality, APSNet also outperforms SampleNet in terms of inference time.

\begin{table}[h]
\begin{center}
\caption{Inference time comparison of three sampling methods with different sample size $m$. The time is reported in millisecond. $^*$ denotes the best default recommended by each paper.}\label{table:time}
\begin{tabular}{ c r r r r r}
\hline

\hline
  $m$ &          32  &          128 &     256 &     512   \\
\hline
SampleNet-G     &    7.63 & 7.54& 7.79&7.94          \\
SampleNet-M$^*$     &    44.33 & 135.23 & 261.47 &515.30        \\
APSNet-G$^*$        &    9.21  & 12.84 & 17.68 & 27.48      \\
APSNet-M        &    45.91  & 139.83 & 269.40 & 525.38      \\
\hline

\hline
\end{tabular}
\end{center}
\vspace{-20pt}
\end{table}

\subsection{Registration}

The task of registration aims to align two point clouds by predicting rigid transformations (e.g., rotation and translation) between them. To save memory and computation power, the registration is conducted on the key points that are sampled from the original point clouds. We follow the work of PCRNet~\cite{sarode2019pcrnet} to construct a point cloud registration network (the task network), and train PCRNet on the point clouds of 1,024 points of the Car category from ModelNet40. Following the settings in SampleNet, 4,925 pairs of source and template point clouds are generated for training, where a template is rotated by three random Euler angles in the range of \([-45\degree, 45\degree]\) to obtain the source. An additional 100 source-template pairs are generated from the test split for evaluation. The mean rotation error (MRE) between the predicted rotations and ground-truth rotations is used as the evaluation metric.

Table~\ref{table:registration} reports the performances of five sampling methods for registration. Similar to the results on classification and reconstruction, APSNet outperforms SampleNet consistently by a notable margin, and achieves the state-of-the-art results in this task. Without leveraging labeled training data, APSNet-KD again demonstrates an impressive performance that is close to supervised APSNet.

\begin{table*}[h]
\centering
%\begin{adjustbox}{width=\columnwidth,center}
\caption{The mean rotation errors of five sampling methods with different sample sizes $m$ on the ModelNet40 dataset for registration. M$^*$ denotes the original results from the SampleNet paper~\cite{lang2020SampleNet}. The lower, the better.}
\scalebox{0.85}{
\begin{tabular}{c|c|c|ccc|cc|cc}\Xhline{5\arrayrulewidth}
%\backslashbox{Methods}{Metrics}
   &\makebox[2.7em]{RS}&\makebox[2.7em]{FPS} &  \multicolumn{3}{c|}{\makebox[2.7 em]{SampleNet}}  & \multicolumn{2}{c|}{\makebox[2.7em]{APSNet}} &  \multicolumn{2}{c}{\makebox[2.7em]{APSNet-KD }}  \\
      $m$ & & &\makebox[2.7em]{G}&\makebox[2.7em]{M}& \makebox[2.7em]{M*}&\makebox[2.7em]{G}&\makebox[2.7em]{M}   &\makebox[2.7em]{G}&\makebox[2.7em]{M} \\\Xhline{2\arrayrulewidth}
8&   63.37&  31.44&  9.72&   8.27&  10.51&  \textbf{5.47} &  9.40& 5.93&   10.51\\
16&  43.89&  20.34&  12.14 & 7.45&  8.21&   \textbf{4.50}&  7.18&  5.01&   7.07\\
32&  27.06&  12.97&  10.81 & 6.13&  5.94&   \textbf{4.37}&  5.82&  4.56&   6.07  \\
64&  16.88&  7.89&   10.93 & 5.38&  5.31&   \textbf{4.42}&  6.34&  4.49&   4.97   \\
% 128&    - &         -&       6.80&        4.24&  5.00&       4.46&   5.12   \\
% 256&    - &         -&       5.00&        4.28&  4.76&       4.26&   4.81  \\
% 512&   73.82&     86.06&    90.18&       89.78&  90.18&    8.69&    -\\
\Xhline{5\arrayrulewidth}
\end{tabular}}
%\end{adjustbox}%
\label{table:registration}
\end{table*}

\vspace{-5pt}
\paragraph{Visualization of Attention Coefficients} 

\begin{figure}[h]
\centerline{
\includegraphics[width = 8cm]{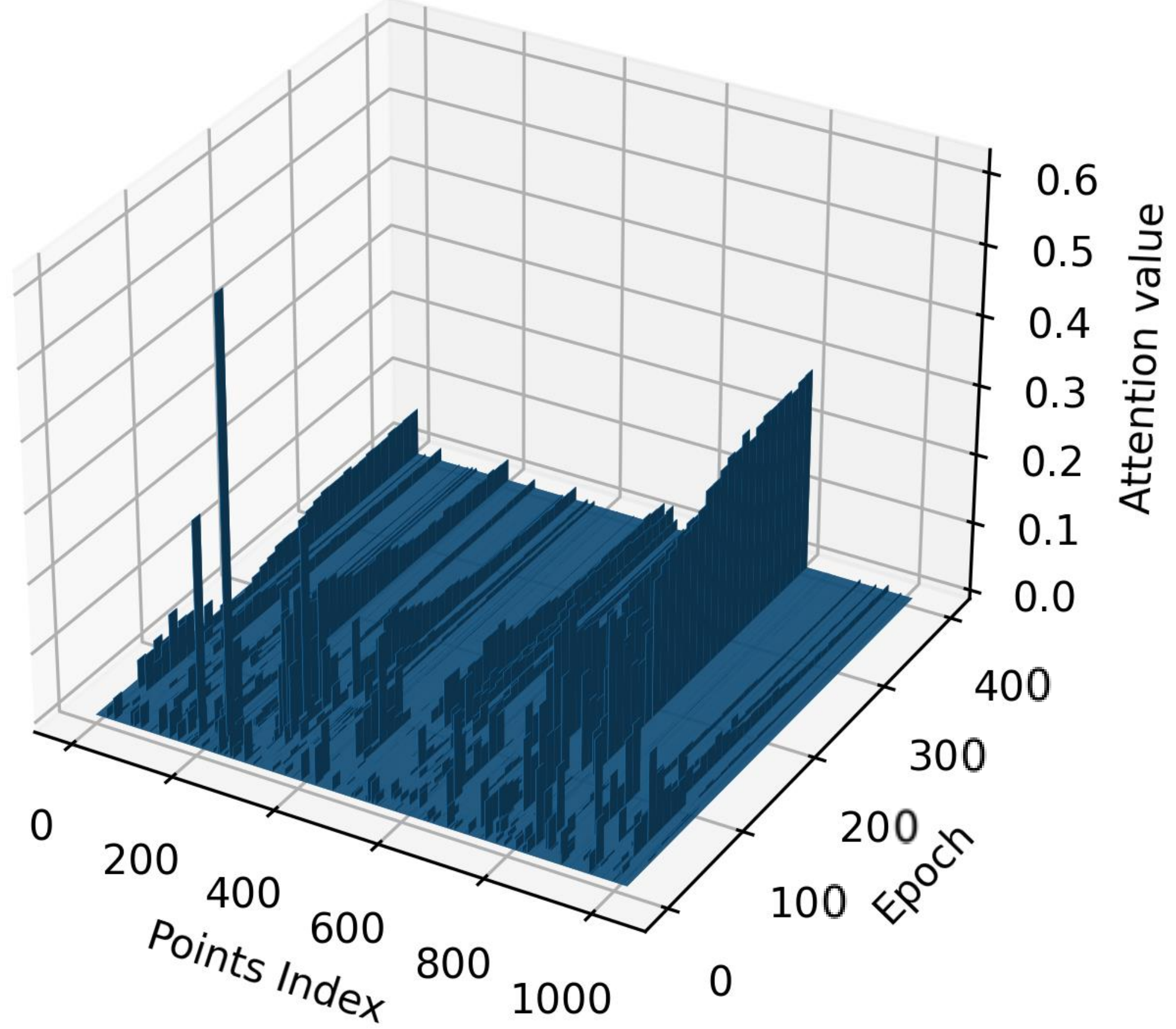}}\vspace{-5pt}
\caption{Evolution of the attention coefficients of APSNet when generating the $t$-th sample. As the training proceeds, the coefficients become sparser with peak values on a few points.}
\label{fig:weight_evo}
\end{figure}

For the task of registration, we further visualize the evolution of attention coefficients during the training process. Specifically, we monitor the attention coefficients Eq.~(3) when generating a point at a specific time step $t$ (the $t$-th sample) from a given point cloud of 1024 points. Figure~\ref{fig:weight_evo} visualizes the evolution of attention coefficients over 400 training epochs. At beginning of the training, the sampler cannot decide which point from the point cloud is the most important one to sample, manifested by the dense cluttered coefficients. As the training proceeds, the attention coefficients become sparser with peak values on 2-3 points. Further, these attention coefficients are stablized in the late training epochs and consistently concentrate on a few the same points, demonstrating the training stability of APSNet.

%We are trying to sample 16 points from this point cloud.
\end{document}